\documentclass[letterpaper, 10 pt, twocolumn, conference, numbers,sort&compress]{ieeeconf} 

\IEEEoverridecommandlockouts                              

\usepackage{cite}
\usepackage{amsmath,amssymb,amsfonts}
\usepackage{algorithmic}
\usepackage{graphicx}
\usepackage{textcomp}
\usepackage{lipsum} 
\usepackage{caption}
\usepackage{xcolor}
\usepackage{booktabs}%
\usepackage{subfigure}
\def\BibTeX{{\rm B\kern-.05em{\sc i\kern-.025em b}\kern-.08em
    T\kern-.1667em\lower.7ex\hbox{E}\kern-.125emX}}

\newcommand{\nop}[1]{} 

\begin{document}

\title{\LARGE \bf  Learning to Change: Choreographing Mixed Traffic Through Lateral Control and Hierarchical Reinforcement Learning \\
}
\author{Dawei Wang$^{1}$, Weizi Li$^{2}$, Lei Zhu$^{3}$, Jia Pan$^{1}$
\thanks{$^{1}$Dawei Wang and Jia Pan are with Department of Computer Science and TransGP at University of Hong Kong, Hong Kong SAR {\tt\small dawei@connect.hku.hk; jpan@cs.hku.hk}}
\thanks{$^{2}$Weizi Li is with Min H. Kao Department of Electrical Engineering and Computer Science at University of Tennessee, Knoxville, TN, USA {\tt\small weizili@utk.edu}}%
\thanks{$^{3}$Lei Zhu is with Department of Industrial and Systems Engineering at University of North Carolina at Charlotte, Charlotte, NC, USA {\tt\small lei.zhu@charlotte.edu}}%
}


\maketitle


\begin{abstract}
The management of mixed traffic that consists of robot vehicles (RVs) and human-driven vehicles (HVs) at complex intersections presents a multifaceted challenge. 
Traditional signal controls often struggle to adapt to dynamic traffic conditions and heterogeneous vehicle types. 
Recent advancements have turned to strategies based on reinforcement learning (RL), leveraging its model-free nature, real-time operation, and generalizability over different scenarios.
We introduce a hierarchical RL framework to manage mixed traffic through precise longitudinal and lateral control of RVs. 
Our proposed hierarchical framework combines the state-of-the-art mixed traffic control algorithm as a high level decision maker to improve the performance and robustness of the whole system. 
Our experiments demonstrate that the framework can reduce the average waiting time by up to 54\% compared to the state-of-the-art mixed traffic control method.
When the RV penetration rate exceeds 60\%, our technique consistently outperforms conventional traffic signal control programs in terms of the average waiting time for all vehicles at the intersection.
\end{abstract}

\section{Introduction}  
With continuous advancement of autonomous driving technology, the efficient management of mixed traffic that consists of robot vehicles (RVs) and human-driven vehicles (HVs) holds paramount significance for our future transportation system. 
Effective control strategies can alleviate congestion, reduce travel times, and enhance overall traffic flow, thereby enhancing the efficiency and productivity of urban transportation networks.
Examining our current road networks, intersections serve as pivotal points where diverse streams of traffic converge, necessitating efficient management to ensure smooth and safe passage for all traversing vehicles. 
Mixed traffic control through RVs represents a frontier in traffic engineering innovation, offering promising avenues to enhance traffic management over road networks including intersections.

\begin{figure}[ht]
\centerline{\includegraphics[width=\linewidth]{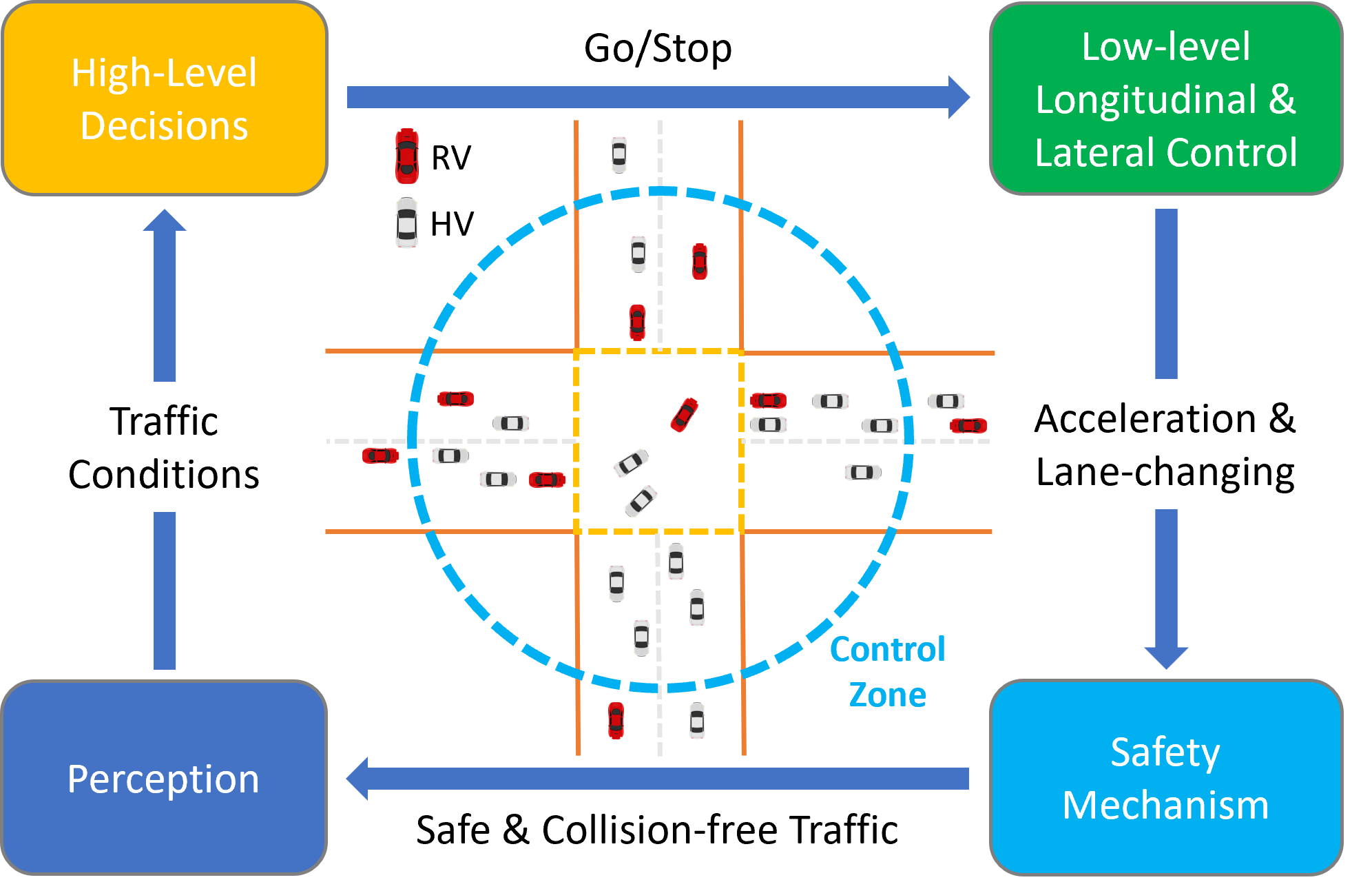}} 
\caption{\small{Our framework starts with a perception system gathering both macroscopic and microscopic traffic conditions. 
Next, high-level decisions, Go/Stop, are made for the RVs. 
Subsequently, the framework generates low-level longitudinal and lateral control commands for the RVs.  
Lastly, a safety mechanism is deployed to resolve conflicting traffic streams and prevent vehicle collisions, ensuring safety in mixed traffic control at complex intersections.} 
}
\label{fig:overview}
\vspace{-1.5em}
\end{figure}

Recent research has underscored the efficacy of RL in orchestrating mixed traffic across a variety of scenarios, including ring roads and figure-eight configurations~\cite{wu2021flow, wei2019mixed}, highway bottlenecks and merge points~\cite{vinitsky2018lagrangian, feng2021intelligent}, two-way intersections~\cite{yan2021reinforcement}, and roundabouts~\cite{jang2019simulation}. Furthermore, alternative investigations delve into the utilization of image-based observations, as opposed to precise metrics like position and velocity, for optimizing mixed traffic flow~\cite{Villarreal2023Pixel}, as well as the examination of how real-world human driving behaviors may perturb established mixed traffic control strategies~\cite{Poudel2024Simulated}.

Considering mixed traffic control at large-scale complex intersections, state-of-the-art approach~\cite{wang2023learning} has predominantly entailed the generation of high-level directives, primarily in the form of binary Go and Stop decisions for the RVs.
However, this strategy is demonstrably inadequate given the increasingly intricate dynamics inherent to contemporary traffic environments. 
Consequently, there is a strong demand to expand the operative action space to incorporate comprehensive longitudinal and lateral controls, enabling fine-grain traffic regulation. 
Such a pivotal expansion not only fosters enhanced maneuverability and adaptability within mixed traffic but also lays the foundation for realizing more sophisticated and responsive mixed traffic control methods.

Our goal is to enhance the existing mixed traffic control methodology by broadening the action space to encompass full longitudinal and lateral control capabilities. To achieve robust performance, we introduce a hierarchical framework that integrates the established high-level decision-making processes of prior state-of-the-art mixed traffic control methods with a novel reinforcement learning (RL) policy generating longitudinal and lateral actions for the RVs. 
Additionally, we implement a safety mechanism within the framework to mitigate conflicts and preempt collision risks within intersections. 
An overview of our framework is provided in Fig.~\ref{fig:overview}.
Validated through extensive experiments under real-world traffic settings, our framework showcases a remarkable capability, slashing the average waiting time by up to 54\% compared to the state-of-the-art mixed traffic control method by Wang et al.~\cite{wang2023learning}. 
Notably, when the penetration rate of RVs exceeds 60\%, our method surpasses traditional traffic signal control in minimizing the average waiting time for all vehicles at intersections. 
To the best of our knowledge, our framework is the first to achieve comprehensive control, encompassing high-level decisions along with low-level longitudinal and lateral maneuvers, within large-scale mixed traffic control. 
This comprehensive control enables the coordination of hundreds of vehicles traversing unsignalized intersections with superior efficiency.

\begin{figure*}[ht]
\centerline{\includegraphics[width=\linewidth]{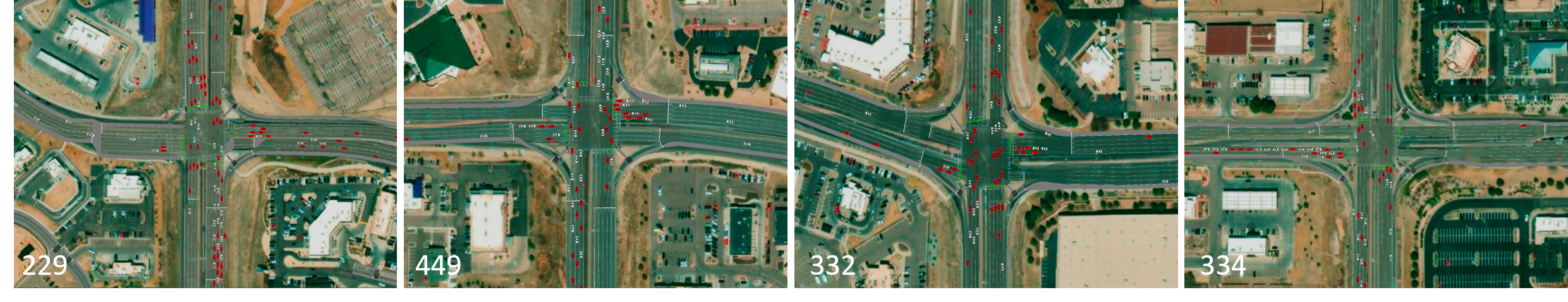}}
\caption{\small{Mixed traffic control at four real-world intersections situated in Colorado Springs, CO, USA, using actual traffic data sourced directly from these intersections. RVs are in red and HVs are in white. The RV penetration rate is 50\%.
Our framework enables efficient traffic flows at these intersections without the presence of traffic lights. 
}} 
\label{fig:traffic_sim_real}
\vspace{-1.5em}
\end{figure*}
\section{Related Work}
We will first introduce studies regarding mixed traffic control and then discuss previous work on the longitudinal and lateral planning of robot vehicles (RVs). 

\subsection{Mixed Traffic Control}
Conventional methods for  mixed traffic control often hinge on mathematical modeling, such as formulating the task as an optimization problem~\cite{wang2023general, wang2019controllability, karimi2020cooperative}. For example, Yang et al.~\cite{yang2020intelligent} develop an optimization strategy to manage mixed traffic flow at unsignalized intersections. Zhao et al.~\cite{zhao2018optimal} address the challenge by optimizing traffic at roundabouts. 
While effective, conventional methods usually impose artificial assumptions about traffic flow or fail to encapsulate heterogeneous traffic behaviors. Additionally, traditional approaches often struggle to scale and generalize across different scenarios.

In response, recent investigations have explored the efficacy of reinforcement learning (RL) as a viable alternative, leveraging its capacity to navigate the intricate behaviors inherent to mixed traffic environments without being encumbered by the same set of assumptions regarding traffic flow dynamics. 
Recent studies have showcased the promise of RL-based mixed traffic control across various scenarios, encompassing ring roads, figure-eight configurations, highway bottlenecks, merge points, two-way intersections,  roundabouts, and larger road networks~\cite{wu2021flow,wei2019mixed,vinitsky2018lagrangian,feng2021intelligent, jang2019simulation, yan2021reinforcement,Wang2024Privacy,Villarreal2024Eco}. Furthermore, emerging research delves into alternative approaches, such as employing image-based observations for control~\cite{Villarreal2023Pixel}, investigating the impact of real-world human driving behaviors on control strategies~\cite{Poudel2024Simulated}, and using large language models (LLMs) to design reward function for mixed traffic control~\cite{Villarreal2023Can}. 
Recently, Wang et al.~\cite{wang2023learning} present an RL-based method for controlling large-scale mixed traffic in real-world traffic settings.  
However, their mixed traffic control algorithm only generates high-level decisions (Stop or Go) for the RVs. 
A complete learning-based RV controller for large-scale mixed traffic remains blank.

\subsection{Longitudinal and Lateral Planning of RVs}
The decision-making and planning of RVs can be mainly divided into three categories: sequential planning, behavior-aware planning, and end-to-end planning~\cite{schwarting2018planning}. To provide some examples, Wang et al.~\cite{wang2018reinforcement} introduce an RL-based method for vehicle lane-changing maneuvers.
Following high-level decision-making, low-level planners are utilized to generate feasible driving trajectories~\cite{Shen2022IRL,Li2019ADAPS}. 
Recently, an Multi-Agent RL (MARL) algorithm is developed to address traffic on-ramp merging by enhancing safety enhancements through a priority-based supervisor~\cite{chen2023deep}. 
As another example, an MARL strategy~\cite{dong2020drl} is developed to improve collaborative sensing by integrating Graph Convolutional Network (GCN)~\cite{kipf2016semi} and Deep Q-Network (DQN)~\cite{mnih2013playing}. 
The study shows promising outcomes in a simulated environment featuring a three-lane freeway with two off-ramps. 
While these approaches offer effective and safe planning solutions for RVs, none of them has demonstrated the capability to scale to large-scale mixed traffic scenarios that involve many conflicting traffic streams, thus limiting their applicability.

\section{Methodology}
We give an overview of our framework followed by explaining its components in detail.

\subsection{Overview} 
The flowchart of our framework is shown in Fig.~\ref{fig:overview}, commencing with the perception system of the RVs, which retrieves observations for processing. These observations include metrics such as the average waiting time of vehicles in a queue and the length of the queue of a traveling direction. 
Next, the high-level decisions of each RV, i.e., Go or Stop, are made by an RL policy. 
These decisions serve as recommendations for each RV, signaling whether it is advisable to enter or refrain from entering an intersection based on the current traffic condition. 
Following the high-level decisions, the low-level planner 
formulates longitudinal and lateral control commands, including accelerations and lane-changing decisions, to facilitate the navigation of the RVs passing through the intersection. 
Finally, a safety mechanism is implemented to preempt potential collisions and conflicts within the intersection, ensuring the integrity of the framework and the safety of all vehicles involved.

\subsection{Intersectional Traffic Flow}

At a standard four-way intersection, traffic movement is characterized by eastbound (E), westbound (W), northbound (N), and southbound (S), with three distinct turning options left (L), right (R), and cross (C). 
For instance, the notation E-L signifies left-turning traffic traveling eastbound, while E-C represents crossing traffic also moving eastbound. We define a `conflict' as the convergence of two moving directions, such as E-C and N-C. In total, we identify eight traffic streams that are prone to potential conflicts: E-L, E-C, W-L, W-C, N-L, N-C, S-L, and S-C. 
In contrast, the \textit{conflict-free} set $\mathcal{C}$ is set to be \{(S-C, N-C), (W-C, E-C), (S-L, N-L), (E-L, W-L), (S-C, S-L), (E-C, E-L), (N-C, N-L), (W-C, W-L)\}; any movement pair that is not  in $\mathcal{C}$ may lead to conflicts.

In our formulation, we exclude consideration of right-turning traffic based on the observation that, in our test environment, the majority of intersections are equipped with dedicated right-turn lanes.
Consequently, right-turning vehicles either bypass the intersection entirely or occupy it only fleetingly. Moreover, traffic regulations in many countries, such as the U.S., often exempt right-turn vehicles from waiting for a green light, further mitigating the need to coordinate their movements with traffic from other directions. 
Our experiments validate that this exclusion minimally impacts intersection traffic control and coordination.

To improve the applicability of our framework for mixed traffic control, we employ actual traffic data to reconstruct traffic patterns, subsequently subjecting them to high-fidelity simulations. We simulate mixed traffic conditions by randomly designating each spawned vehicle to be either RV or HV according to a pre-specified RV penetration rate.

\subsection{High-level Control Decisions} 
We adopt the state-of-the-art mixed traffic control algorithm from Wang et al.~\cite{wang2023learning} as the high-level decision maker in our hierarchical RL framework. 
The high-level action space is $A = \{  \text{Go}, \text{Stop} \}.$ The action $a_i^t \in A$ of an RV indicates whether vehicle $i$ should proceed into the intersection or come to a halt at the entrance of the intersection. 
The high-level observation space, on the other hand, of RV $i$ at $t$ is 
\begin{equation}
    o^t_i = \oplus_{j}^J \langle {l}^{t,j}, {w}^{t,j} \rangle \oplus_{j}^J \langle {m}^{t,j} \rangle \oplus \langle d^t_i \rangle,
\end{equation}
where $\oplus$ is the concatenation operator and $J = 8$ is the total number of traffic moving directions. Here, $d^t_i$ denotes the distance from RV $i$'s current position to the intersection. 
We additionally track the queue length $l^{t,j}$, the average waiting time $w^{t,j}$, and the occupancy map $m^{t,j}$ for each of the eight traffic moving directions within the intersection.


\subsection{Low-level Longitudinal and Lateral Control}
Upon receiving high-level decisions, there are two scenarios: `Go' indicating the recommended entry of the RV into the intersection and `Stop' indicating the recommended halt decision for the RV.
In response to `Go', we apply maximum acceleration to the RV; when encountering `Stop', the RL-based longitudinal and lateral control policy generates appropriate acceleration and lane changes for the RV.

We formulate the low-level longitudinal and lateral control of RV as a Partially Observable Markov
Decision Process (POMDP) which consists 7-tuples: $(\mathcal{S}, \mathcal{A}, \mathcal{T}, \mathcal{R}, \pi_{\theta}, \mathcal{V}_{\phi}, \mathcal{D} )$, where $\mathcal{S}$ represents the set of all possible states, $\mathcal{A}$ represents the set of all possible actions, $\mathcal{T}$ represents the probability distribution over next states given the current state and action, $\mathcal{R}$ represents the immediate reward received after taking an action in a state, $\pi_{\theta}$ represents the parameterized stochastic policy that maps states to probability distributions over actions, $\mathcal{V}_{\phi}$ represents the parameterized function that estimates the expected return from a given state under the current policy, $\mathcal{D}$ represents the collection of past experiences \{(state, action, reward, next state, done flag)\} for training the policy and value function.

Deep neural networks are used to represent both the policy and value function in our formulation. In particular, Proximal Policy Optimization (PPO)~\cite{schulman2017proximal} is employed to optimize the policy. 
The training loss comprises both the policy loss and the value function loss: 
\vspace{-.2em}
\[ L(\theta) = L^{PPO}(\theta) + L^{VF}(\theta). \]

\noindent The policy loss is formulated as
\vspace{-.2em}
\[ L^{PPO}(\theta) = \mathbb{E}_t \left[ \min \left( r_t(\theta) \hat{A}_t, \text{clip}(r_t(\theta), 1 - \epsilon, 1 + \epsilon) \hat{A}_t \right) \right], \] 

\noindent where \( r_t(\theta) = \frac{\pi_{\theta}(a_t|s_t)}{\pi_{\theta_{\text{old}}}(a_t|s_t)} \) is the ratio between the current policy \( \pi_{\theta}(a_t|s_t) \) and the old policy \( \pi_{\theta_{\text{old}}}(a_t|s_t) \), \( \hat{A}_t \) is the advantage estimate, and \( \epsilon \) is a hyperparameter controlling the policy update clipping and set to $0.3$ empirically. 
The value function loss is given by
\vspace{-.2em}
\[ L^{VF}(\theta) = \mathbb{E}_t \left[ \left( V_{\theta}(s_t) - V_{\text{target}}(s_t) \right)^2 \right], \]

\noindent where \( V_{\theta}(s_t) \) is the value predicted by the value function under the current policy, and \( V_{\text{target}}(s_t) \) is the target value.


\begin{table*}[t] 
    \centering
    \begin{tabular}{ccccccccc}
        \toprule
        & \multicolumn{8}{c}{Reduced Average Waiting Time (\%)}   \\  
        \cmidrule(l){2-9} 
        & \multicolumn{4}{c}{Compared to Wang~\cite{wang2023learning}}  & \multicolumn{4}{c}{Compared to TL} \\        
        \cmidrule(l){2-5} \cmidrule(l){6-9}     
        Intersection 229             & 18.98\% & 6.66\% & * & * & * & * & 0.51\%  & 30.94\% \\ 
        Intersection 449             & 31.65\% & 28.23\% & 5.74\% & * & 79.94\% & 76.71\%  & 78.76\%   & 69.71\%\\ 
        Intersection 332             & 18.48\%  & 11.73\%  & 35.85\% & 26.59\% & 3.94\% & 11.73\%  &20.06\%   & 41.58\%\\ 
        Intersection 334             & 38.73\% & 54.67\% & 48.32\% & 45.82\% & 17.77\%  & 39.44\%  & 58.34\% & 47.82\%\\
        \midrule
        RV Penetration Rate             & 40\% & 50\% & 60\% & 70\% & 40\% & 50\% & 60\% & 70\%   \\ 
        \bottomrule
    \end{tabular}
    \caption{\small{Reduced average waiting time in percentage at four intersections under various RV penetration rates---calculated as $(ours - baseline)/baseline$. 
    Our method achieves comparable or superior performance to the state-of-the-art mixed traffic control algorithm by Wang et al.~\cite{wang2023learning}.
    Our method outperforms traffic signals when the RV penetration rate is 40\% or higher in general. * indicates no improvement over the baseline methods.  } }
    \label{tab:intersection_overall}
    \vspace{-1.5em}
\end{table*}

\textbf{Observation Space}: The low-level controller's observation comprises both macroscopic and microscopic observations. 
The macroscopic observation includes
\begin{equation}
    oa^t_i = \oplus_{j}^J \langle {l}^{t,j}, {w}^{t,j} \rangle \oplus_{j}^J \langle {m}^{t,j} \rangle,
\end{equation}
which mirrors the structure of the high-level decision maker's observation.
The microscopic observation consists
\begin{equation}
    oi^t_i = \oplus \langle d^t_i \rangle \oplus \langle {cl}^t_i \rangle \oplus \langle {cr}^t_i \rangle,
\end{equation}
where $d^t_i$ denotes the distance from RV $i$'s current position to the intersection, while ${cl}^t_i$ and ${cr}^t_i$ respectively indicate whether the left lane and right lane are controlled by the RV. 
For instance, ${cl}^t_i$ is set to 1 if at least one RV is detected in the left lane of the ego vehicle, otherwise ${cl}^t_i$ is set to 0.

\textbf{Action Space}: The action space is continuous and comprises of longitudinal acceleration ($acc$) and lane changing probability ($lc$), both ranging from -1 to 1. 
If $lc < -0.33$, a left lane-changing action is initiated; if $lc > 0.33$, a right lane-changing command is issued; if $lc$ falls within $[-0.33, 0.33]$, the RV will stay in its current lane.

\textbf{Reward}: 
The reward function aims to minimize the average waiting time of all vehicles approaching the intersection:
\begin{equation}
    r(s^t, a^t, s^{t+1}) = -\sum w_i, 
\end{equation}
where $w_i$ represents the normalized waiting time of all vehicles within the intersection control zone (see Fig.~\ref{fig:overview}).

\subsection{Safety Mechanism}
Managing a multi-agent system through RL poses many challenges~\cite{lowe2017multi}. 
A pivotal and open challenge is the development of a multi-agent autonomous system that is not only effective but also provably safe~\cite{gu2022review}. 
Therefore, we implement a safety mechanism to post-process actions generated by the hierarchical RL policy. This ensures safety by resolving potential conflicts within the intersection.

The safety mechanism comprises several speed limit zones contingent on the distance to the entrance of the intersection. 
Should a vehicle exceed the specified speed limit within these zones, automatic braking is activated to decelerate the vehicle. 
The speed limits are defined as
\begin{equation} 
\begin{cases}
  3~m/s,  & \ \text{if } 20~m <d^t_i<=30~m ;\\
  2~m/s,  & \ \text{if } 10~m <d^t_i<=20~m ;\\
  1~m/s,  & \ \text{if } 5~m <d^t_i<=10~m ;\\
  0~m/s,  & \ \text{if } d^t_i<=5~m. \\
\end{cases}
\label{eq:speed_limit}
\end{equation}

Apart from managing conflicts within the intersection, addressing the risk of collisions between vehicles is also crucial. 
We delegate this responsibility to the underlying traffic simulator SUMO~\cite{lopez2018microscopic}: 
if an RV is on a collision trajectory, either longitudinally or laterally with other vehicles, SUMO will inhibit the RV from executing the control commands from the RL policy and instead activate an emergency braking maneuver to prevent the collision.

\section{Experiments and Results}
We begin by introducing mixed traffic simulation for evaluation. Following that, we present the baseline methods employed and outline the evaluation metric. 
Finally, we present the overall results and provide a detailed analysis of the effectiveness of RV lane-changing behaviors.

\subsection{Mixed Traffic Simulation}
To enable the interaction between RVs and HVs under real-world traffic settings, it is imperative to reconstruct traffic patterns using real-world traffic data before proceeding with high-fidelity simulations. 
We reconstruct traffic dynamics via turning count data provided by the city of Colorado Springs, CO, USA\footnote{https://coloradosprings.gov/}.
The data record vehicle movements as well as the digital maps. 
Vehicles in simulation are routed using jtcrouter\footnote{https://sumo.dlr.de/docs/jtrrouter.html} based on the turning count data. 
Upon entering the simulation, each spawned vehicle will be randomly assigned as either an RV or HV, determined by the RV penetration rate.
HVs use Intelligent Driver Model (IDM)~\cite{treiber2000congested} to calculate longitudinal acceleration. 
RVs, on the other hand, use our hierarchical RL framework to decide high-level decisions and low-level controls, including longitudinal acceleration and lateral lane changing.
Examples are shown in Fig.~\ref{fig:traffic_sim_real}.



\subsection{Baselines and Evaluation Metric}
We evaluate our proposed method using two baselines: \textbf{TL}: the traffic signal program deployed in the city of Colorado Spring, CO, and \textbf{Wang}~\cite{wang2023learning}: the state-of-the-art mixed traffic control method at intersections. 
The metric we employ for evaluation is the average waiting time. 
The waiting time for each vehicle is defined as the cumulative time it spends stationary within the control zone after entering it. 
The average waiting time for an intersection is determined as the mean waiting time for all vehicles present at that intersection and inside the control zone.


\begin{figure*}[ht]
\centerline{\includegraphics[width=\linewidth]{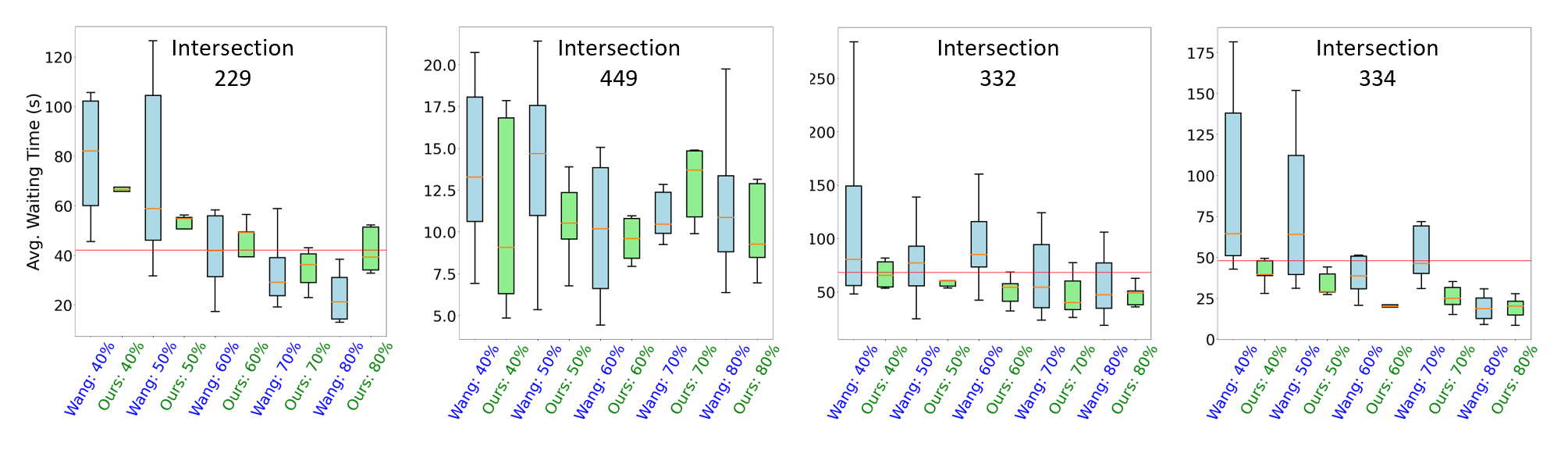}}
\vspace{-1em} 
\caption{
The overall results measured in average waiting time at four intersections between our technique and Wang et al.~\cite{wang2023learning}. 
The red line represents the average waiting time of traffic light control baseline (TL). 
Our method consistently outperforms the TL baseline when RV penetration rate reaches 60\% or higher. 
Furthermore, in the majority of scenarios, our approach exhibits reduced average waiting times compared to Wang under the same RV penetration rates. 
For intersection 449, it is worth noting that the TL baseline has a high value of 45 seconds, thus it is excluded from the plot.
This indicates that, at intersection 449, both our technique and Wang outperform TL under all tested RV penetration rates starting at 40\%. 
For both intersection 332 and 334. our method consistently outperforms TL and Wang when the RV penetration rate $\ge 40\%$. 
In general, our technique also demonstrates much lower variance than Wang's, showing improved robustness and performance in mixed traffic control as a result of incorporating lateral and longitudinal control for the RVs.
}
\label{fig:box}
\vspace{-1.5em}
\end{figure*}

\subsection{Intersection Performance}
We report the results of our hierarchical RL framework at four real-world intersections shown in Fig.~\ref{fig:traffic_sim_real}. 
All training is conducted using the Intel i9-13900K processor and NVIDIA RTX 4090 graphics card.

Table~\ref{tab:intersection_overall} presents a comparison of the results obtained using our method and the baseline methods. 
The findings demonstrate a decrease in average waiting time across all four intersections and various RV penetration rates. 
Our approach consistently matches or surpasses the performance of the latest mixed traffic control algorithm by Wang et al.~\cite{wang2023learning}. Our method also demonstrates superior performance compared to traffic signal control (TL), particularly when the RV penetration rate exceeds 40\%. An asterisk (*) denotes cases where no improvement over the baseline methods is observed.
The reasons for this phenomenon is the following. 
Compared to prior research~\cite{wang2023learning}, when the RV penetration rate is sufficiently high, the need for lateral control diminishes as the majority of traffic in the network comprises RVs, leading to natural lane regulation. Consequently, the performance enhancement achieved by mitigating lane irregularities is not substantial as the RV penetration rate increases. 




\textbf{Intersection 229}. 
In Fig.~\ref{fig:box}, we show our method's performance at intersection 229.
The impact of varying RV penetration rates on the average waiting time is analyzed.
Overall, the average waiting time consistently decreases as the RV penetration rate increases from 40\% to 80\%, shown in both Wang et al.~\cite{wang2023learning} and our proposed method. In most scenarios, our method demonstrates better performance compared to Wang under equivalent RV penetration rates. 
However, occasional exceptions are observed due to varying traffic conditions and training inconsistencies. 
Similar to Wang, our proposed approach surpasses the TL baseline when the RV penetration rate exceeds 70\%. 
Furthermore, our method showcases reduced evaluation variance, indicating not only superior performance over the state-of-the-art mixed traffic control algorithm~\cite{wang2023learning} but also enhanced stability across evaluation scenarios.

\textbf{Intersection 449}.  
The TL baseline at this intersection is much higher than other intersections, i.e., 45 seconds. 
So, we exclude it from the plot. 
In other words, at this intersection, both our technique and Wang significantly outperform the TL baseline when the RV penetration rate $\ge 40\%$. 
Moreover, our framework exhibits superior performance compared to Wang when the RV penetration rates are 40\%, 50\%, 60\%, and 80\%.

\textbf{Intersection 332}. 
The results depicted in Fig.~\ref{fig:box} illustrate that across all RV penetration rates, our framework consistently outperforms the baseline method by Wang. It is worth noting that, while Wang manages to surpass the TL baseline with a minimum RV penetration rate of 70\%, our technique also demonstrates superiority over the TL baseline in general, starting from an RV penetration rate 40\%.

\textbf{Intersection 334}. 
The results in Fig.~\ref{fig:box} highlight the superior performance of our method over the TL baseline and Wang across all RV penetration rates.
While Wang starts to surpass the TL baseline when the RV penetration rate exceeds 70\%, our method achieves this phenomenon starting at 40\% RV penetration rate. Additionally, our method achieves much lower variance compared to Wang.

\begin{figure*}[ht]
\centerline{\includegraphics[width=\linewidth]{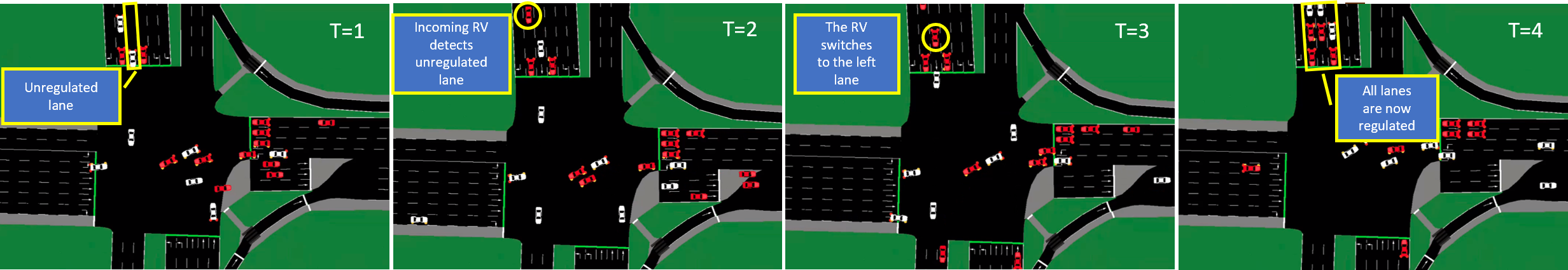}}
\caption{\small{Lane-changing behavior for mixed traffic control. 
An example shows an RV detecting an unregulated lane and subsequently switching to it in order to regulate it.  
Initially, with limited RV presence, HVs can exploit unregulated lanes, causing potential risk to the traffic within the intersection. 
This risk is resolved when an incoming RV detects it and decides to switch to the unregulated lane.  
As a result, a coordinated RV fleet forms to regulate southbound lanes, effectively mitigating conflicts and enhancing traffic efficiency within the intersection.
}}
\label{fig:lc_demo}
\vspace{-1.5em}
\end{figure*}

\subsection{Analysis of Lane-changing Behaviors} 

Previous research by Wang et al.~\cite{wang2023learning} suggest a critical issue: when the RV penetration rate is low, managing mixed traffic through RVs becomes difficult. 
This is due to an influx of HVs will enter the intersection causing gridlocks.  
While the fundamental reason is the shortage of control sources, i.e., limited RVs due to a low penetration rate, the absence of lateral control could also contribute to the issue -- the RV cannot regulate traffic in nearby lanes by pursuing strategical lane changing. 
Our framework bridges this gap, enabling more precise control of traffic flows at the intersection.

An example illustrating the effectiveness of our approach is depicted in Fig.~\ref{fig:lc_demo}.
Initially, only two RVs traveling southbound are positioned in front of the entrance line of the intersection, regulating the traffic on the two lanes.
In the absence of RVs, other lanes heading southbound remain unregulated (one of them is highlighted in yellow).
These lanes are at risk of HVs unexpectedly entering the intersection, thereby increasing the likelihood of collisions and conflicts within it. 
Subsequently, an incoming RV approaches the intersection and detects the existence of unregulated lanes.
Recognizing the potential risk, the RV controlled by our framework initiates lane changing and switches to the left (unregulated) lane.
With the presence of an RV, the lane is now regulated.
Essentially, the RV fleet collaboratively acts as traffic light control, strategically obstructing all lanes to prevent both HVs and additional RVs from entering the intersection. 
As a result, it effectively coordinates the traffic flow within the intersection, improving traffic efficiency.

\begin{figure}[htbp]
\centerline{\includegraphics[width=\linewidth]{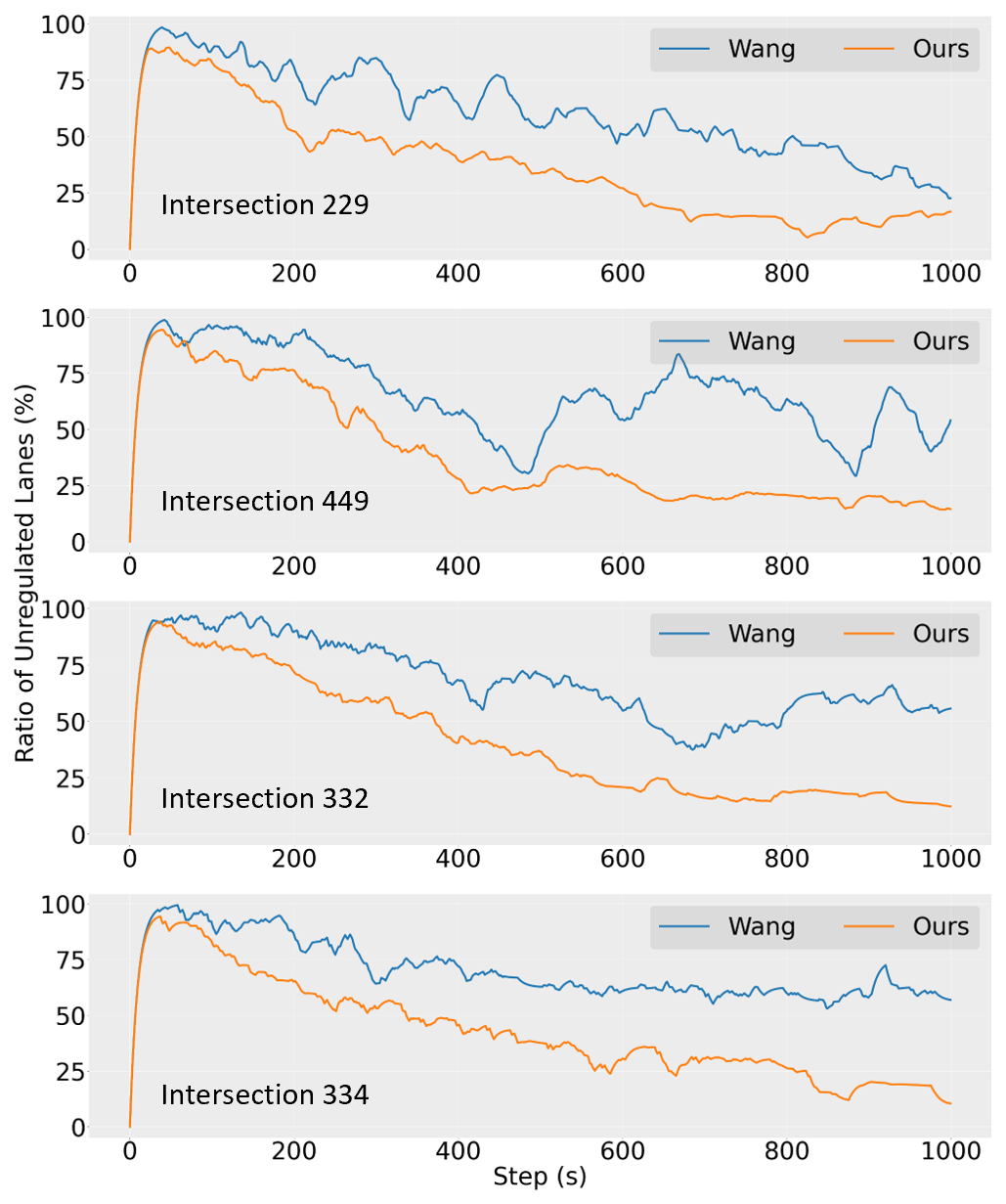}}
\vspace{-.5em}
\caption{\small{Comparison of the ratio of unregulated lanes between our method and Wang et al.~\cite{wang2023learning}. 
The RV penetration rate is 40\%. 
The blue line denotes the percentage of unregulated lanes when our method is employed, while the orange line represents Wang's performance. 
Our method results in a faster reduction in unregulated lanes due to proactively lane-changing behaviors.}} 
\label{fig:failure_control}
\vspace{-.5em}
\end{figure}

We show more systematic analysis of RV lane-changing behaviors on regulating traffic in Fig.~\ref{fig:failure_control}.  
One prominent factor contributing to the suboptimal performance observed in prior research by Wang et al.~\cite{wang2023learning} is the absence of lane-changing behaviors of RVs in regulating traffic. 
To study the effectiveness of the addition of lane-changing behaviors, we conduct an assessment to quantify the ratio of unregulated lanes during our experiments.  
In Fig.~\ref{fig:failure_control}, the blue line represents the percentage of unregulated lanes when our method is engaged, while the orange line signifies the performance of Wang's approach~\cite{wang2023learning}. 
Initially, both methods exhibit nearly 100\% unregulated lanes due to the absence of vehicles in the beginning of the simulation. 
Subsequently, as vehicles ingress the network, this percentage drops accordingly. Notably, our framework accelerates the reduction compared to Wang. 
This expedited decline is attributed to our method's proactive lateral and longitudinal control over RVs, facilitating lane changes and prioritizing the regulation of unregulated lanes wherever feasible.


\section{Conclusion and Future Work}

We present a novel hierarchical reinforcement learning framework for enhancing mixed traffic control at intersections by integrating high-level decisions and low-level longitudinal and lateral maneuvers. 
Additional safety mechanism is implemented to ensure the integrity of the framework and the safety of all vehicles crossing the intersection.
We conduct extensive experiments to validate our framework under real-world traffic settings. 
Our approach demonstrates significant improvements in reducing the average waiting time of mixed traffic (up to 54\%) compared to state-of-the-art mixed traffic control methods, particularly in scenarios with high penetration rates of robot vehicles. 
Furthermore, our approach outperforms conventional traffic signal control in reducing the average waiting time for all vehicles at intersections once the penetration rate of RVs surpasses 60\%.
To the best of our knowledge, our framework is the first to comprehensively control and coordinate large-scale mixed traffic traversing unsignalized intersections with superior efficiency.
The outcomes offer valuable insights for the advancement of future intersection traffic management systems with enhanced efficiency and adaptability. 

There are many future research directions.  
First, we plan to test our framework on heterogeneous traffic, including trucks, buses, and motorcycles. 
This will enable us to assess the framework's functionality and adaptability across a diverse range of vehicle types and traffic scenarios.
Second, we want to conduct emission analysis and pursue potential emission reductions within mixed traffic.  
Finally, we would like to extend our work to cover a larger urban area by integrating the framework with existing large-scale traffic modeling, estimation, and simulation techniques~\cite{Wilkie2015Virtual,Li2017CityFlowRecon,Lin2022GCGRNN,Guo2024Simulation}.


\bibliographystyle{IEEEtran}
\bibliography{bibliography}

\end{document}